\documentclass[letterpaper]{article} 
\usepackage{aaai2026}  
\usepackage{times}  
\usepackage{helvet}  
\usepackage{courier}  
\usepackage[hyphens]{url}  
\usepackage{graphicx} 
\usepackage{natbib}  
\usepackage{caption} 
\usepackage{algorithm,algorithmicx, algpseudocode}
\usepackage{amsmath}
\usepackage{multirow}
\usepackage{mathrsfs}
\usepackage{amsfonts}
\usepackage{diagbox}
\usepackage{arydshln}

\usepackage{newfloat}
\usepackage{listings}
\DeclareCaptionStyle{ruled}{labelfont=normalfont,labelsep=colon,strut=off} 
\floatstyle{ruled}
\newfloat{listing}{tb}{lst}{}
\floatname{listing}{Listing}
\title{Deferred Poisoning: Making the Model More Vulnerable via Hessian Singularization}
\author {
    Yuhao He\textsuperscript{\rm 1},
    Jinyu Tian\textsuperscript{\rm 1}\thanks{corresponding author},
    Xianwei Zheng\textsuperscript{\rm 2}, 
    Li Dong\textsuperscript{\rm 3}, 
    Yuanman Li\textsuperscript{\rm 4}, 
    Jiantao Zhou\textsuperscript{\rm 5}
}
\affiliations {
    \textsuperscript{\rm 1}Faculty of Innovation Engineering, Macau University of Science and Technology\\
    \textsuperscript{\rm 2}School of Mathematics,
Foshan University\\
    \textsuperscript{\rm 3}Faculty of
Electrical Engineering and Computer Science, Ningbo University\\
    \textsuperscript{\rm 4}College of Electronics and Information Engineering, Shenzhen University\\
    \textsuperscript{\rm 5}	Department of Computer and Information Science, University of Macau\\
    3250004430@student.must.edu.mo,
    jytian@must.edu.mo,
    alex.w.zheng@hotmail.com,
    dongli@nbu.edu.cn,
    yuanmanli@szu.edu.cn,
    jtzhou@um.edu.mo

}

\usepackage{bibentry}

\begin{document}

\maketitle

\begin{abstract}
Recent studies have shown that deep learning models are very vulnerable to poisoning attacks. Many defense methods have been proposed to address this issue. However, traditional poisoning attacks are not as threatening as commonly believed. This is because they often cause differences in how the model performs on the training set compared to the validation set. Such inconsistency can alert defenders that their data has been poisoned, allowing them to take the necessary defensive actions. In this paper, we introduce a more threatening type of poisoning attack called the \textbf{Deferred Poisoning Attack}. This new attack allows the model to function normally during the training and validation phases but makes it very sensitive to evasion attacks or even natural noise. We achieve this by ensuring the poisoned model's loss function has a similar value as a normally trained model at each input sample but with a large local curvature. A similar model loss ensures that there is no obvious inconsistency between the training and validation accuracy, demonstrating high stealthiness. On the other hand, the large curvature implies that a small perturbation may cause a significant increase in model loss, leading to substantial performance degradation, which reflects a worse robustness. We fulfill this purpose by making the model have singular Hessian information at the optimal point via our proposed Singularization Regularization term. We have conducted both theoretical and empirical analyses of the proposed method and validated its effectiveness through experiments on image classification tasks. Furthermore, we have confirmed the hazards of this form of poisoning attack under more general scenarios using natural noise, offering a new perspective for research in the field of security.
\end{abstract}

\begin{links}
    \link{Code}{https://github.com/Anson-He/DPA}
\end{links}

\section{Introduction}

Deep learning models have achieved remarkable achievements in fields involving Computer Vision \cite{resnet} to Natural Language Processing \cite{transformer}. Their success is largely attributed to the proliferation of large-scale datasets, such as ImageNet \cite{imagenet}, COCO \cite{coco}. Nevertheless, recent studies have highlighted the potential hazards posed by poisoning attacks which can undermine the integrity of deep learning models by introducing malicious data into training datasets \cite{better-safe-than-sorry,fowl2021adversarial,tao2022can}.

Most prevalent poisoning attacks \cite{NEURIPS2018_22722a34, EM, REM, Fang2019LocalMP} share a common limitation: their malicious intentions are highly apparent. The victim may notice a significant disparity in the performance of the model between the training and validation datasets. In response, strategies such as adversarial training \cite{bai2021recent}, appropriate data preprocessing methods \cite{apbench}, or the elimination of anomalous \cite{liao2018defense} data can be employed as defensive measures against these attacks.

\begin{figure*}[t]
    \centering
    \includegraphics[width=0.8\textwidth]{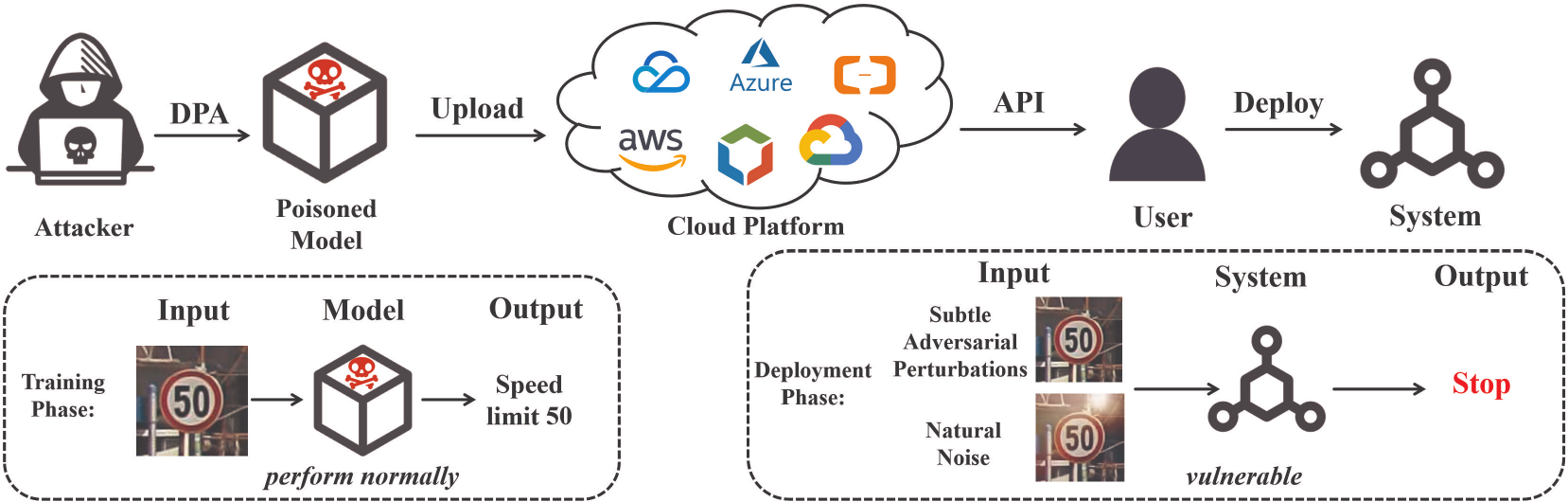}
    \caption{The scenario considered by DPA.}
    \label{fig:MLaaS}
\end{figure*}

In this paper, we reveal a more threatening method of poisoning attack namely \textbf{Deferred Poisoning Attack (DPA)}. As the term "\textit{Deferred}" implies, this attack does not disrupt the training process, allowing the model to maintain normal performance over the validation set. Instead, the "\textit{toxicity}" of the attack manifests by undermining the model's robustness at the deployment stage. Fig. \ref{fig:MLaaS} illustrates the scenario of our attack by the case of Machine Learning as a Service (MLaaS) \cite{ribeiro2015mlaas}. Attackers can use DPA to train a model that performs normally during training and testing, correctly identifying signs like "Speed Limit 50" to gain user trust. However, once deployed in systems like autonomous driving, this vulnerable model becomes susceptible to evasion attacks or even natural disturbances such as fog, rain, and lighting variations. This can cause critical failures, such as misclassifying a speed limit sign as "STOP".

\begin{figure}[t]
    \centering
    \includegraphics[width=0.4\textwidth]{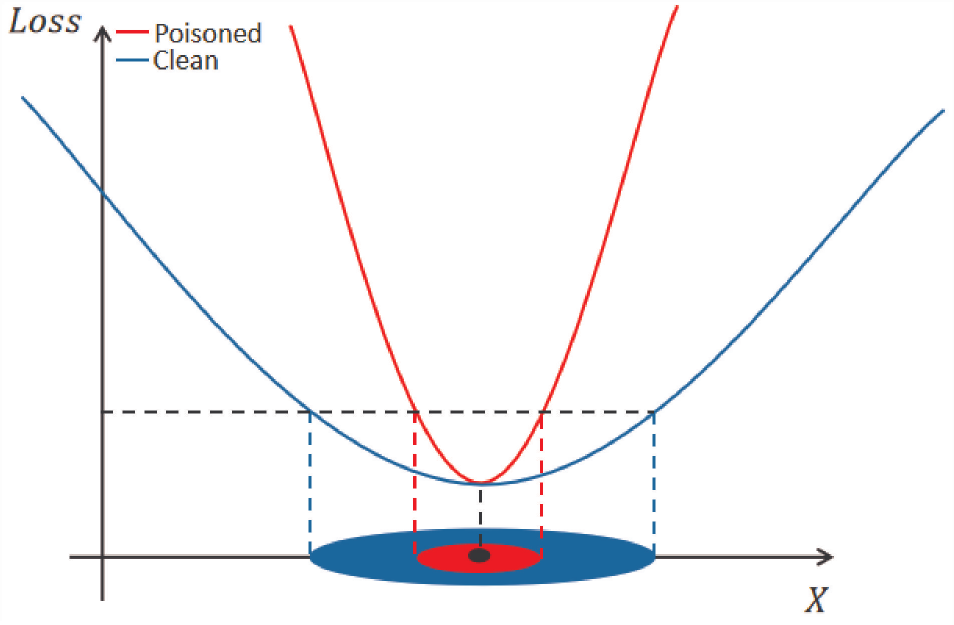}
    \caption{The illustration of the motivation of DPA.}
    \label{fig:manuscript}
\end{figure}

We fulfill the purpose of DPA by forcing the model trained over the contaminated dataset to converge to a similar point as the one trained over the clean dataset, making the poisoned model perform normally on the validation dataset. On the other hand, we enlarge the local curvature of the poisoned model around each sample in the training dataset to amplify the sensitivity of the poisoned model. Fig. \ref{fig:manuscript} illustrates the above motivation. A large local curvature (the red curve) results in a significant increase in model loss with a small perturbation of a given sample. In contrast, a small local curvature (the blue curve) enables the model loss to remain stable even with a large perturbation. Formally, a large local curvature implies that the Hessian matrix is ill-conditioned with a large conditional number \cite{boyd2004convex}. Along this line, our DPA generates poisoned samples to induce the model trained on this contaminated dataset to become singularization (a large conditional number.) with respect to the input samples. 

In summary, our principal contributions are as follows:

\begin{itemize}
    \item The proposed DPA, to the best of our knowledge, has not been previously addressed in the literature, thus revealing a new threat within the field of artificial intelligence.
    \item We propose a novel regularization term to amplify the local curvature of the poisoned model that generates noise patterns exhibiting both visual stealthiness and adversarial effectiveness.
    \item Compared to traditional data poisoning methods, DPA incurs a significantly lower attack cost (subtle perturbation) while demonstrating superior transferability and robustness.
    \item We validate the generality of the DPA across a broader range of scenarios.
\end{itemize}

\section{Related work}
\label{Related Work}
Poisoning attacks aim to insert malicious samples into training datasets, thereby compromising the ability of the resulting models \citep{biggio2012poisoning,Biggio2017WildPT,Jagielski2018ManipulatingML,Liu2018TrojaningAO}. \citet{EM} first proposed a method for calculating perturbations by minimizing empirical risk, which impedes the models' ability to extract valuable information from images. \citet{REM} made a more robust improvement by minimizing adversarial training loss to obtain a more robust set of unlearnable examples. However, previous poisoning attacks directly compromised the integrity of the models, making it easy for us to detect anomalies in the model's performance on the validation set during the training phase, which then allows us to take defensive measures. In this paper, we identify a form of deferred poisoning that appears normal during the model training phase yet renders the poisoned model extremely vulnerable. This kind of stealthiness can mislead researchers, thereby posing a serious threat to their work.

\section{Deferred poisoning attack}
In this section, we introduce the principles of DPA, detailing how to generate such perturbations and providing both the necessary theoretical and empirical analysis.
\subsection{Problem statement}
\paragraph{Assumptions on adversaries’s capability} We assume that adversaries have access to the entire training dataset in the MLaaS scenario but are unaware of the target model's parameter structure or any output, and do not interfere with the target model's training process.
\paragraph{Objectives} We discuss this issue in the context of image classification. Assuming a classic K-class task.Let $x\in{\mathcal{X}}\subset{\mathbb{R}^{d}}$ represents the training samples, and $y\in{\mathcal{Y}}=\{1,\ldots,K\}$ are the labels, $ \hat{x}=x+\delta$ represents the poisoned samples, and $\delta\in\Delta\subset{\mathbb{R}^{d}}$ is the perturbation emphasized in this paper. The perturbation $\delta$ is constrained by ${\Vert \delta \Vert}_{p}\leq\epsilon$, where ${\Vert \cdot \Vert}_{p}$ is the $L_p$ norm, and $\epsilon$ is typically set to a relatively small number to ensure the perturbation is imperceptible to the human. 

As we discussed in the introduction, the objective of our DPA is to ensure that the model trained on a contaminated dataset performs normally on a clean dataset, but remains highly sensitive to noise, including adversarial perturbations and even natural noises. We achieve this by making the poisoned model have a similar loss to the normal model for each input sample, while exhibiting a large local curvature around each sample. We resort to Hessian Singularization to amplify this local curvature. The overall objective function is as follows:

\begin{align}
\label{eq:vulnerability}
\mathop{\arg\min}_{\theta,\delta} \bigg[\mathcal{L}(f_{\theta}({x}),y)&+\mathcal{L}(f_{\theta}(\hat{x}),y)-Q(f_{\theta}( \hat{x}),y)\bigg]\\
\nonumber &\text{s.t.}\;{\Vert \delta \Vert}_{\infty}\leq\epsilon.
\end{align}

In the objective function (\ref{eq:vulnerability}), the first term, the cross-entropy loss, is designed to ensure the poisoned model performs normally on clean examples. The second and third terms are dedicated to updating the poisoning perturbation $\delta$. Notably, our approach differs significantly from existing poisoning attack methods \cite{wu2022one,tian2022comprehensive}, which typically aim to increase the model loss on poisoned examples so as to degrade the model's performance on the validation dataset. Our proposed loss still allows the model to converge on the poisoned examples but introduces a unique design: the Hessian Singularization term $Q(f_{\theta}( \hat{x}),y) =\textit{tr}(H^{T}H)$, where $H$ is the Hessian matrix of the loss function with respect to the input $\hat{x}$. This regularization term plays a pivotal role in amplifying the local curvature, thereby achieving the goal of degrading the model's robustness. The value of $Q(f_{\theta}( \hat{x}),y)$ is positively related to the curvature around the sample $\hat{x}$. More of the relevant theoretical derivation about the Hessian Singularization term will be discussed in Session \ref{Theoretical Analysis}. In the upcoming section, we will first introduce how to solve the optimization problem (\ref{eq:vulnerability}) and then present the entire scheme of our DPA.

\subsection{Generating perturbation}

\begin{figure*}[t]
    \centering
    \includegraphics[width=0.73\textwidth]{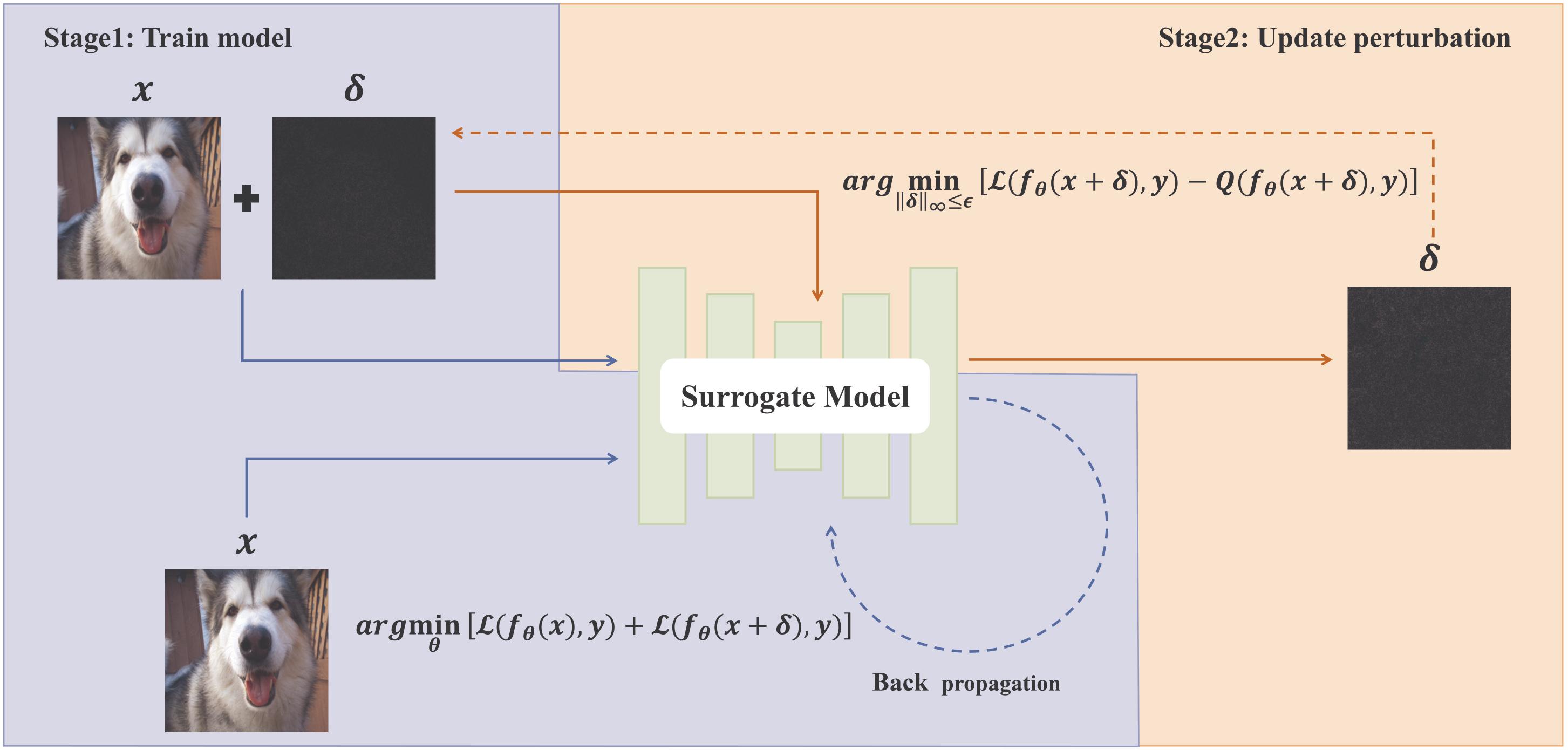}
    \caption{The framework of DPA.}
    \label{fig:fraework}
\end{figure*}

Empirically, to ensure that the optimization of the problem (\ref{eq:vulnerability}) converges effectively, we divide this problem into two alternating iterative sub-optimization problems, which have been widely adopted in the research area of poisoning attack \cite{munoz2017towards, REM}.
Specifically, as shown in Fig. \ref{fig:fraework}, the first phase is the model training stage. We consider the following sub-problem
\begin{equation}
\label{eq:vulnerability-1}
\theta \leftarrow \mathop{\arg\min}_{\theta} \bigg[\mathcal{L}(f_{\theta}({x}),y)+\mathcal{L}(f_{\theta}(\hat{x}),y) \bigg].
\end{equation} 
This problem ensures that the model converges on the poisoned samples while still functioning effectively on clean samples. The second phase is the perturbation update stage via solving the following sub-problem, 
\begin{equation}
\label{eq:vulnerability-2}
\delta \leftarrow \mathop{\arg\min}_{{\Vert \delta \Vert}_{\infty}\leq\epsilon} \bigg[\mathcal{L}(f_{\theta}(\hat{x}),y)-Q(f_{\theta}( \hat{x}),y)\bigg]
\end{equation}
where the model $f_{\theta}$ is derived from the first phase and all the parameters $\theta$ are fixed in this stage.  Upon solving the second sub-problem (\ref{eq:vulnerability-2}), the resulting perturbation will force the poisoned model updated in the next round of stage 1 to exhibit a large local curvature around each input sample. These two stages will be iteratively repeated until convergence. More details about our DPA are provided below.

The procedure of the proposed DPA is depicted in Algorithm \ref{Alg}. The outer optimization is used for training the victim model (row 5 to row 6), calculating the sum of the cross-entropy loss $\mathcal{L}$ for the model on both original and poisoned data for each minibatch to update model parameters at each iteration. The inner optimization is for updating the perturbation (row 10 to row 11), calculating the difference between the cross-entropy loss $\mathcal{L}$ on the poisoned data and $Q$ for each minibatch to update perturbation in each iteration. Here $Q(f_{\theta}^{*}(x+\delta),y) = \textit{tr}(H^{T}H)$ where $H$ is the Hessian matrix of the cross-entropy loss function $\mathcal{L}$ with respect to $x+\delta$. 

\begin{algorithm}[t]
    \centering
    \renewcommand{\algorithmicrequire}{\textbf{Input:}}
    \renewcommand{\algorithmicensure}{\textbf{Output:}}
    \caption{Algorithm for generating perturbation.}
    \label{Alg}
    \begin{algorithmic}[1]
        \Require Training set $D_{tr}$; Victim model $f_{\theta}$; Model learning rate $\eta_{\theta}$; Perturbation learning rate $\eta_{\delta}$; Epoch of the model training process $I_{\theta}$; Number of the perturbation adapting iterations $I_{\delta}$ ; 
        \Ensure Perturbation set $\Delta$; 

        \State Initialize model parameter $\theta$;
        \State Initialize $\Delta$ as zero matrices.;
        \For {$i=1,2,\cdots,I_{\theta}$}
            \For {Minibatch $B\subset D_{tr}$ and Minibatch $P\subset \Delta$}
                \State
                $g_{\theta}\leftarrow \mathbb{E}_{(x,y)\in B,\delta\in P}[\nabla_{\theta}(\mathcal{L}(f_{\theta}(x,y))+\mathcal{L}(f_{\theta}(x+\delta,y)))]$
                \State
                $\theta \leftarrow \theta-\eta_{\theta}g_{\theta}$
            \EndFor
            \For {$j=1,2,\cdots,I_{\delta}$}
                 \For {Minibatch $B\subset D_{tr}$ and Minibatch $P\subset \Delta$}
                    \State
                    $g_{P}\leftarrow \mathbb{E}_{(x,y)\in B,\delta\in P}[\nabla_{(x+\delta)}(\mathcal{L}(f_{\theta}(x+\delta,y))-Q(f_{\theta}(x+\delta,y)))]$
                    \State
                    $P \leftarrow P-\eta_{\delta}g_{P}$
                \EndFor
            \EndFor
        \EndFor
    
        \State \Return $\Delta$.
    \end{algorithmic}
\end{algorithm}

\subsection{Hessian singularization}
\label{Theoretical Analysis}
In this section, we will explain why the Hessian Singularization could impact the robustness of deep models.
Consider the loss function of a model denoted by $l$ and a given input $x$ and assume that this function is strictly convex around a small neighbor of the $x$. For any small perturbation $h$ around $x$, the loss function satisfies the following conditions \cite{boyd2004convex}:
\begin{equation} 
\label{eq:rearranged}
\begin{aligned}
\nabla l(x)^{T}(h)+\frac{\sigma_{min}(H)}{2}\|h\|_{2}^{2} \leq l(x+h) - l(x) \leq \\
\nabla l(x)^{T}(h)+\frac{\sigma_{max}(H)}{2}\|h\|_{2}^{2}
\end{aligned}
\end{equation}
where $\sigma_{min}(H)$ and $\sigma_{max}(H)$ denote the smallest and largest singular values of the Hessian matrix, respectively. 

The inequations (\ref{eq:rearranged}) delineate the range of variation of the loss function in the vicinity of a given example when subjected to minor perturbations. The greater the disparity between $\sigma_{min}(H)$ and $\sigma_{max}(H)$, the larger the variation range of the loss function, thereby facilitating adversaries in identifying a perturbation that induces a significant error or even for some natural perturbations. Motivate by this conclusion, we thus design DPA to generate perturbations such that the poisoned dataset can significantly increase the maximum singular value $\sigma_{max}(H)$ of the hessian matrix $H$ of the loss function at each sample during the model training process. However, the computation of $\sigma_{max}(H)$ typically requires methods such as Singular Value Decomposition, which generally is differentiable. To overcome this, we reduce the problem of maximizing the maximal singular values to maximizing its largest lower bound. That is 
\begin{equation}
\label{eq:max_sigma}
\sigma_{\max}(H)\geq [\frac{1}{n}tr(H^{T}H)]^{\frac{1}{2}}.
\end{equation}
where $tr()$ represents the trace of a matrix (see \cite{lancaster1985theory} for the details of this inequality). Consequently, this leads to the proposed Hessian Singularization regularization term $Q(f_{\theta}^{*}(x+\delta),y)=\text{tr}(H^{T}H)$ in the objective function (\ref{eq:vulnerability}). 

\begin{figure}[t]
    \centering
    \includegraphics[width=0.4\textwidth]{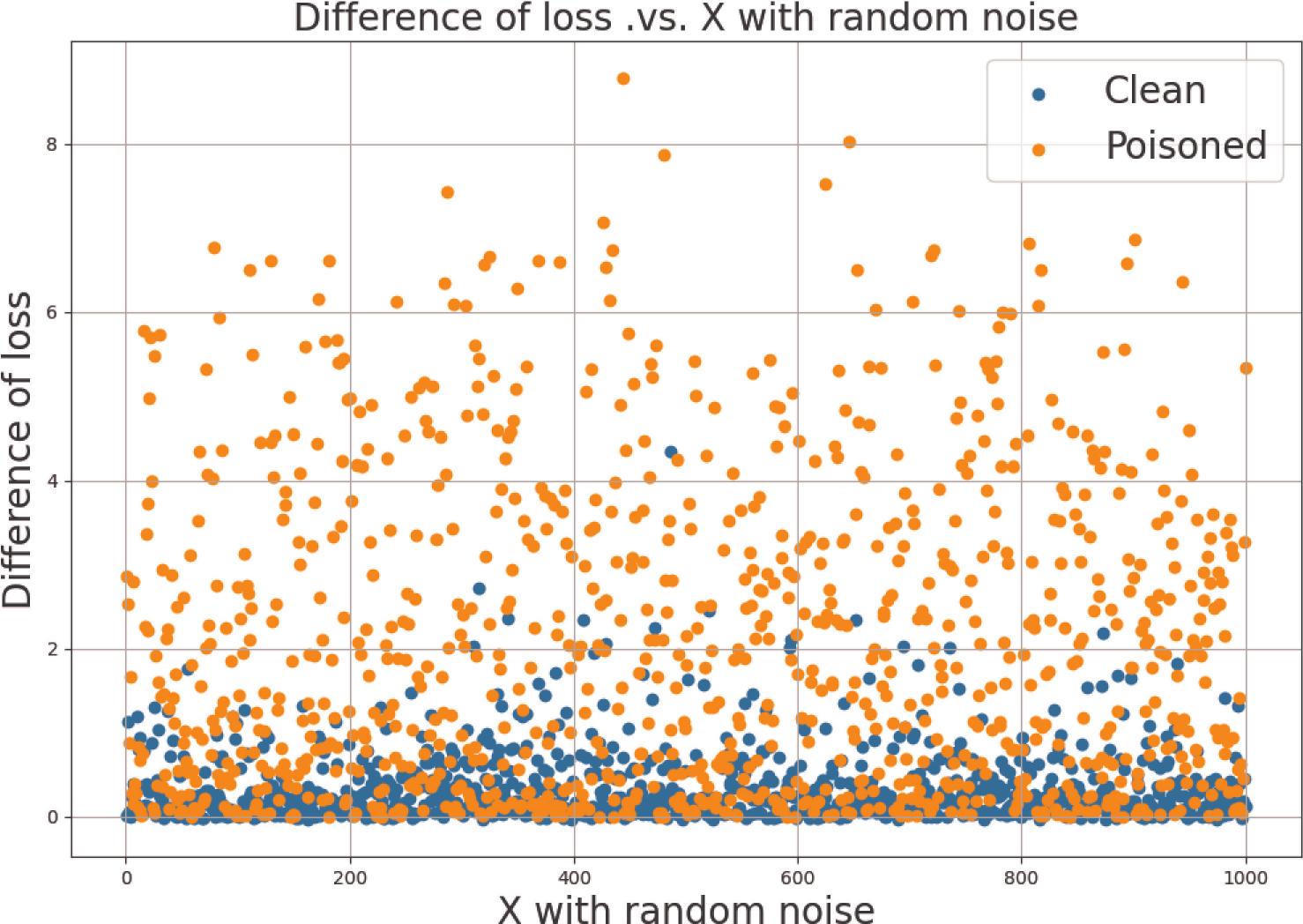}
    \caption{Comparison of the loss values of the poisoned model and the clean
model to random noise.}
    \label{fig:empirical}
\end{figure}

Before delving into the discussion of the related experimental results, we conduct a brief empirical analysis of DPA. As shown in Fig. \ref{fig:empirical}, the x-axis represents the application of 1000 instances of Gaussian noise to a fixed image, and the y-axis represents the difference in the model's loss value for the noisy image compared to the loss value for the clean image. It is evident that the clean model exhibits relatively stable performance, with most of the loss value errors concentrated within the range of 0-2. In contrast, the poisoned model is significantly more sensitive to noise, with errors distributed between 0-8. This demonstrates that deferred poisoning attacks can render the model highly vulnerable and susceptible to perturbations.

\subsection{Relaxation of Hessian singularization term}
\label{hvp}
Actually, calculating the Hessian matrix is computationally expensive. Suppose the dimensionality of the input is $p$, determining the Hessian matrix requires computing the Jacobian matrix of the gradient function, which involves $p^2$ backpropagation steps. In practice, on an NVIDIA 3090 ti, computing the Hessian matrix takes about 8 seconds for images in CIFAR10. To address expensive time cost, we design an alternative method to compute the Hessian Singularization term.

Recalling that $tr(H^{T}H)=\|H\|^{2}_{F}$, given an arbitrary unit vector $v$, we have $\|Hv\|_{2} \leq \|H\|_{F}\|v\|_{2}=\|H\|_F$. Thus, we can relax the problem of the maximization of the term $tr(H^{T}H)=\|H\|^{2}_{F}$ to the maximization of the tight lower bound $\|Hv\|_{2}^{2}$ which is the Hessian Vector Product and has a quick computation algorithm supported by Pytorch function \textit{torch.autograd.functional.hvp}.

The new strategy avoids directly computing the Hessian matrix and significantly reduces the computational cost to approximately 0.008 seconds and still maintains the effectiveness of our DPA.

\begin{table*}[t]
\centering
\begin{tabular}{c|cccc|cccc|cccc}
\hline
& ACC    & $\hat{\rho}_{F}\downarrow$    & $\hat{\rho}_{P}\downarrow$    & $\hat{\rho}_{D}\downarrow$  & ACC   & $\hat{\rho}_{F}\downarrow$  & $\hat{\rho}_{P}\downarrow$    & $\hat{\rho}_{D}\downarrow$ & ACC     & $\hat{\rho}_{F}\downarrow$    & $\hat{\rho}_{P}\downarrow$    & $\hat{\rho}_{D}\downarrow$ \\ \hline
Dataset-Model  & \multicolumn{4}{c|}{TinyImageNet-DesNet121}& \multicolumn{4}{c|}{CIFAR10-VGG16}    & \multicolumn{4}{c}{SVHN-VGG16}     \\ \hline
Clean                    &0.74&0.75&0.48&0.24  &0.89&1.86&0.94&1.19&0.95&3.83&1.98&5.39       \\
Poisoned                 &0.73&\textbf{0.58}&\textbf{0.42}&\textbf{0.08}&0.75&\textbf{0.58}&\textbf{0.50}&\textbf{0.32}&0.88&\textbf{1.54}&\textbf{0.93}&\textbf{1.10}\\ \hline
Dataset-Model  & \multicolumn{4}{c|}{TinyImageNet-ResNet18} & \multicolumn{4}{c|}{CIFAR10-ResNet18} & \multicolumn{4}{c}{SVHN-ResNet18}  \\ \hline
Clean                    &0.72&0.74&0.50&0.28&0.81&1.62&1.01&1.24&0.94&3.33&1.73&2.57\\
Poisoned                 &0.72&\textbf{0.48}&\textbf{0.42}&\textbf{0.09}&0.76&\textbf{0.79}&\textbf{0.57}&\textbf{0.39}&0.89&\textbf{1.78}&\textbf{1.03}&\textbf{1.24}\\ \hline
Dataset-Model  & \multicolumn{4}{c|}{TinyImageNet-ResNet50} & \multicolumn{4}{c|}{CIFAR10-ResNet50} & \multicolumn{4}{c}{SVHN-ResNet50}   \\ \hline
Clean                    &0.73&0.82&0.49&0.23&0.78&2.84&1.48&1.21&0.92&3.24&1.59&2.48\\
Poisoned                 &0.72&\textbf{0.50}&\textbf{0.43}&\textbf{0.07}&0.73&\textbf{1.38}&\textbf{0.62}&\textbf{0.45}&0.88&\textbf{1.63}&\textbf{0.97}&\textbf{1.16}\\ \hline
\end{tabular}%
\caption{The comparison of the accuracy and robustness(\%) of the clean model and poisoned model against different kinds of evasion attacks.}
\label{tab:vulnerable}
\end{table*}

\begin{table*}[t]
\centering
\small
\setlength{\tabcolsep}{1pt}{
\begin{tabular}{c|c|cccc|cccc}
\hline
\multicolumn{2}{c|}{Dataset}           & \multicolumn{4}{c|}{CIFAR10($\epsilon=\frac{3}{255}/\frac{5}{255}/\frac{8}{255}$)}                             & \multicolumn{4}{c}{SVHN($\epsilon=\frac{3}{255}/\frac{5}{255}/\frac{8}{255}$)} \\ \hline
Model & Method & Clean data     & FGSM$\downarrow$           & PGD$\downarrow$              & CW$\downarrow$               & Clean data     & FGSM$\downarrow$             & PGD$\downarrow$             & CW$\downarrow$    \\ \hline
\multicolumn{1}{c|}{\multirow{4}{*}{VGG16}}    & Clean & 0.89           & 0.65           & 0.71           & 0.66           & 0.95           & 0.84           & 0.88          & 0.84         \\
\multicolumn{1}{c|}{}                          & EM    & 0.87/0.83/0.27 & 0.62/0.49/-- & 0.68/0.54/-- & 0.64/0.44/-- & 0.93/0.93/0.29 & 0.80/0.79/-- & 0.84/0.83/--& 0.79/0.78/-- \\
\multicolumn{1}{c|}{}                          & REM   & 0.87/0.54/0.39 & 0.61/--/-- & 0.67/--/-- & 0.62/--/-- & 0.93/0.93/0.93 & 0.79/0.80/0.79 & 0.83/0.84/0.83& 0.78/0.79/0.79 \\
\multicolumn{1}{c|}{}                          & Ours  & 0.75/0.69/0.67 & \textbf{0.14}/\textbf{0.17}/\textbf{0.20} & \textbf{0.21}/\textbf{0.23}/\textbf{0.28} & \textbf{0.14}/\textbf{0.18}/\textbf{0.21} & 0.88/0.83/0.76 & \textbf{0.45}/\textbf{0.41}/\textbf{0.41} & \textbf{0.53}/\textbf{0.50}/\textbf{0.48}& \textbf{0.43}/\textbf{0.40}/\textbf{0.38} \\ \hline
\multicolumn{1}{c|}{\multirow{4}{*}{ResNet18}} & Clean & 0.81           & 0.60           & 0.65           & 0.60           & 0.94           & 0.79           & 0.83          & 0.79           \\
\multicolumn{1}{c|}{}                          & EM    & 0.77/0.66/0.37 & 0.55/0.27/-- & 0.60/0.31/-- & 0.56/0.22/-- & 0.91/0.44/0.27 & 0.73/--/-- & 0.78/--/--& 0.72/--/--  \\
\multicolumn{1}{c|}{}                          & REM   & 0.78/0.40/0.38 & 0.56/-- /--& 0.61/--/-- & 0.57/--/-- & 0.91/0.17/0.42 & 0.73/--/-- & 0.78/--/--& 0.73/--/--   \\
\multicolumn{1}{c|}{}                          & Ours  & 0.76/0.72/0.70 & \textbf{0.24}/\textbf{0.19}/\textbf{0.18} & \textbf{0.32}/\textbf{0.27}/\textbf{0.25} & \textbf{0.23}/\textbf{0.19}/\textbf{0.17} & 0.89/0.83/0.73 & \textbf{0.54}/\textbf{0.45}/\textbf{0.44} & \textbf{0.61}/\textbf{0.53}/\textbf{0.50}& \textbf{0.51}/\textbf{0.42}/\textbf{0.41}    \\ \hline
\multicolumn{1}{c|}{\multirow{4}{*}{ResNet50}} & Clean & 0.78           & 0.58           & 0.63           & 0.59           & 0.92           & 0.76           & 0.81          & 0.75              \\
\multicolumn{1}{c|}{}                          & EM    & 0.76/0.74/0.32 & 0.55/0.52/-- & 0.60/0.57/-- & 0.56/0.53/-- & 0.90/0.40/0.24 & 0.70/--/-- & 0.76/--/--& 0.70/--/--     \\
\multicolumn{1}{c|}{}                          & REM   & 0.75/0.71/0.47 & 0.54/0.44/-- & 0.60/0.49/-- & 0.56/0.44/-- & 0.89/0.46/0.32 & 0.72/--/-- & 0.76/--/--& 0.72/--/--   \\
\multicolumn{1}{c|}{}                          & Ours  & 0.73/0.71/0.66 & \textbf{0.29}/\textbf{0.17}/\textbf{0.16} & \textbf{0.37}/\textbf{0.24}/\textbf{0.23} & \textbf{0.29}/\textbf{0.16}/\textbf{0.16} & 0.88/0.80/0.74 & \textbf{0.47}/\textbf{0.42}/\textbf{0.37} & \textbf{0.55}/\textbf{0.50}/\textbf{0.44}& \textbf{0.44}/\textbf{0.39}/\textbf{0.34}    \\ \hline
\end{tabular}
}
\caption{The performance of various poisoning attacks at different attack scales.}
\label{tab:compare scale}
\end{table*}

\section{Experiments}
\label{Experiments}
In this section, we verify the effectiveness, transferability, and stealthiness of the poisoning perturbations generated by DPA, and verify the performance of HVP. In addition, we also challenge our method with some defense strategies such as data augmentation methods or limited poisoning percentages. Finally, we propose a new training paradigm for DPA to enhance the robustness of the model and compare it with traditional adversarial training.
\subsection{Experiment setting}
\paragraph{Datasets \& Models}We generate perturbation on the CIFAR10 \cite{krizhevsky2009learning}, SVHN \cite{netzer2011reading}, and TinyImageNet \cite{imagenet} datasets respectively. Subsequently, we use the VGG16 \cite{simonyan2014very}, ResNet18 \cite{resnet}, and ResNet50 \cite{resnet} models to train on the CIFAR10 and SVHN datasets. For the TinyImageNet, we use the ResNet18, ResNet50, and DenseNet121 \cite{huang2017densely} models for training. It should be noted that the TinyImageNet is a subset of the ImageNet dataset.

\paragraph{Evaluation metric} The major claim of our work is that models trained over a dataset poisoned by our DPA would exhibit a lower robustness against adversarial perturbation or even natural noises. Thus we quantize the robustness of a victim model by adopting the measure defined by \cite{moosavi2016deepfool}. That is,

\begin{equation}
\label{eq:robustness}
\hat{\rho}(f) = \frac{1}{\lvert\mathscr{D}\rvert}\sum\limits_{x\in\mathscr{D}}\frac{\Vert \hat{r}(x) \Vert_{p}}{\Vert x \Vert_{p}}
\end{equation}

where $\hat{r}(x)$ is the minimal perturbation for a successful attack, and $\mathscr{D}$ denotes the validation set. $p$ is $2$ (e.g., DeepFool \cite{moosavi2016deepfool}) or $\infty$ (e.g., PGD). A smaller $\hat{\rho}(f)$ indicates a model is more vulnerable.

On the other hand, we emphasize that our DPA could significantly decrease the degradation of the model performance on clean examples, exhibiting the stealthiness of our DPA. To this end, we apply traditional poisoning methods such as EM, REM, LSP \cite{LSP}, and AR \cite{AR}, as well as our proposed method to contaminate datasets \cite{apbench}. We assess the classification accuracy of the poisoned models, the closer the accuracy is to that of the clean models, the more effective the stealth of the poisoning attack method.

\begin{figure*}[ht]
    \centering
    \includegraphics[width=0.7\textwidth]{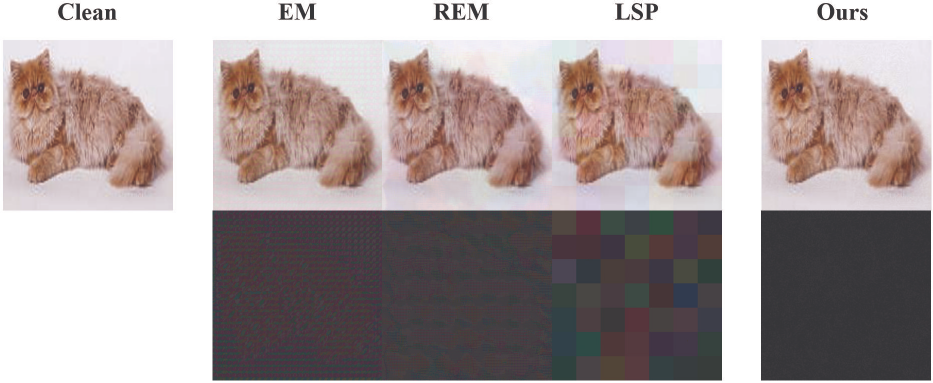}
    \caption{Comparison of examples generated by different Poisoning attacks above the dotted line(i.e. EM, REM and LSP). For each attack, we show the poisoned sample (top) and the magnified (×5) residual (bottom).}
    \label{fig：perturbation_compare}
\end{figure*}

\subsection{Effectiveness of deferred poisoning attack}
\label{Effectiveness}
We claim that the model trained over the dataset poisoned by our DPA could function normally on the validation dataset but is significantly vulnerable to adversarial noise or even natural noise. We support our main claim with the experiment results as follows. 

As shown in Table \ref{tab:vulnerable} where $\hat{\rho}_{F}$, $\hat{\rho}_{P}$ and $\hat{\rho}_{D}$ represent the robustness of victim models attacked by FGSM \cite{FGSM}, PGD \cite{PGD}, and DeepFool respectively, we can observe that the \textit{ACC} of both the clean models and the poisoned models are almost the same. For example, the 6th and 7th rows, corresponding to the 2nd to 5th columns, illustrate the performance of the model trained with ResNet18 on TinyImageNet. After being subjected to DPA, the poisoned model maintains the same accuracy as the clean model on the validation set, both at 0.72. However, the robustness of the poisoned model against FGSM, PGD, and DeepFool adversarial attacks has significantly decreased from 0.74, 0.50, and 0.28 to 0.48, 0.42, and 0.09, respectively. Similarly, by examining the data for other models and datasets in the table, we arrive at the same conclusion.

\begin{table}[t]
\centering
\setlength{\tabcolsep}{0.8mm}{
\begin{tabular}{ccccccc}
\hline
Model &  & Clean & Gaussian & Poisson & Speckle & Rayleigh \\ \hline
\multirow{2}{*}{VGG16} & C    & 0.89  & 0.80     & 0.71    & 0.72    & 0.82     \\
& P & 0.75  & 0.45     & 0.47    & 0.48    & 0.44     \\ \hline
\multirow{2}{*}{ResNet18} & C    & 0.81  & 0.60     & 0.63    & 0.62    & 0.66     \\
& P & 0.76  & 0.41     & 0.49    & 0.41    & 0.43     \\ \hline
\multirow{2}{*}{ResNet50} & C    & 0.78  & 0.63     & 0.61    & 0.65    & 0.68     \\
& P & 0.73  & 0.37     & 0.50    & 0.38    & 0.42     \\ \hline
\end{tabular}%
}
\caption{The impact of natural noise on the poisoned (P) and clean (C) models.}
\label{tab:generalized scenario}
\end{table}

\paragraph{Transferability} It is important to note that, in these examples from Table \ref{tab:vulnerable}, perturbations are generated based on the VGG16 model that was pretrained on the full ImageNet dataset. However, the toxic effects of these perturbations are not only effective on the VGG16 model (white-box scenario) but also transfer to other models such as ResNet18, ResNet50, and DenseNet121 (black-box scenarios). This indicates that DPA possesses excellent transferability, thereby making it highly suitable for black-box attack scenarios. The poisoned model being simultaneously affected by adversarial samples and natural noise indicates that DPA fundamentally reduces the model's robustness to arbitrary perturbations. This corroborates our claim that by increasing the curvature of the loss function around local optima, the model becomes more vulnerable.

\paragraph{Generalized scenario} Adversarial examples do not frequently manifest in real-world scenarios. According to our claim, DPA fundamentally reduces the model's robustness to arbitrary perturbations, that's means poisoned model not only simultaneously affected by adversarial samples but also natural noise. To substantiate this, we introduce random noise sampled from different distributions to the test images to simulate more general conditions. We experimented with natural noise from various distributions: Gaussian Noise, Poisson Noise, Speckle Noise and Rayleigh Noise.


Table \ref{tab:generalized scenario} compares the classification accuracy of clean and poisoned models on CIFAR10 with different sampled random noise injections. It is evident that the poisoned models are significantly more sensitive to random noise, indicating that our method can render the models vulnerable. For example, when perturbing VGG16 with Gaussian noise, the clean model achieves a high classification accuracy of 0.80, remaining virtually unaffected by the noise. However, when the same Gaussian noise is used to disturb the poisoned model, the classification accuracy drops to only 0.45. This indicates that DPA renders the model considerably vulnerable even to natural noise, and poses a significant potential threat to the field of artificial intelligence security.

\subsection{Stealthiness of poisoned samples}
\label{Stealthiness}

In this section, we verify the low attack cost characteristic of DPA and thereby achieve the stealthiness of the poisoned samples. DPA has a significant advantage over traditional data poisoning methods, it requires only a minimal perturbation cost, such as $\epsilon=3/255$, to achieve excellent results. In contrast, data poisoning methods with $l_{\infty}$ norm constraints, such as EM and REM, do not demonstrate effectiveness at this level of perturbation. 

By contrasting the classification accuracy of adversarial samples between the poisoned and clean models, we present the effectiveness of DPA in a more intuitive form. Table \ref{tab:compare scale} presents the classification accuracy of the model against different adversarial samples after being subjected to poisoning attacks with varying perturbation radii, where all adversarial samples' perturbation radii are set to 1/255. It shows that the EM and REM methods do not fully take effect when $\epsilon$ are as low as $3/255$ or even $5/255$ across different models, at least $\epsilon=8/255$ to exhibit aggressiveness. In contrast, our method shows excellent offensiveness even at the scale of $\epsilon=3/255$. For instance, in rows 3 to 6 of Table \ref{tab:compare scale} for the VGG16 model trained on CIFAR10, when $\epsilon=3/255$, the models trained with EM and REM maintain high classification accuracy and accuracy post-adversarial attack. However, the model poisoned by DPA exhibits extremely low accuracy post-adversarial attack, indicating that the models are very vulnerable and can be significantly impacted by adversarial examples due to the approach mentioned. This also suggests that the attack cost of DPA is much lower than that of traditional poisoning attacks, while also ensuring the stealthiness of our method.


As depicted in Fig. \ref{fig：perturbation_compare}, the perturbation we propose is difficult to discern with the naked eye. Compared to other poisoning attacks, even when the residual images between the poisoned and clean images are magnified five times, the noise is hardly noticeable. This illustrates the profound visual stealthiness inherent in our method, which is alarming in terms of its potential for misuse.

\subsection{Comparison with backdoor attacks}
\label{backdoor}
Models attacked by backdoor attacks behave normally during the training phase, but make mistakes when encounter trigges. However, DPA does not implant specific Trojans, more generally, its goal is to make poisoned models particularly vulnerable.

\begin{table}[t]
\centering
\begin{tabular}{cccccc}
\hline
Methods & ACC & ASR & $\hat{\rho}_{F}\downarrow$&  $\hat{\rho}_{P}\downarrow$ & $\hat{\rho}_{D}\downarrow$  \\
\hline
Clean & 0.89 & - &1.86&0.94&1.19      \\
    				
BadNet & 0.83 & 0.95 & 1.51& 0.78 & 0.80\\
    				
Input-aware & 0.84& 0.87 & 1.57& 0.84 & 0.92 \\
    				
LC & 0.85& 0.90 & 1.41&0.76 & 0.74 \\
    				
WaNet & 0.84& 0.58 & 2.29& 0.78& 0.64 \\

SAPA & 0.87 & 0.77 & 1.34 & 0.85 & 1.01 \\
                
\hdashline

Ours & 0.75 & - &\textbf{0.58}&\textbf{0.50}&\textbf{0.32} \\
\hline
\end{tabular}
\caption{The accuracy and robustness(\%) of VGG16 after training CIFAR10 with backdoors implanted.}
\label{tab:backdoor}
\end{table}

Table \ref{tab:backdoor} demonstrates that even when the attack success rate (ASR) is very high, backdoor attacks do not render the model more vulnerable. This indicates that the aggressiveness of DPA is insidious and unique, it can make the model more vulnerable while ensuring normal model performance, bringing unforeseen potential harm to machine learning.

\subsection{Robustness analysis}
\label{Robustness}
We conduct various data augmentation techniques \cite{cutout,image_process} on data-infused with deferred poison. By observing the performance of models trained on these augmented datasets, we assess the robustness of DPA.

Table \ref{tab:defense performence} displays the performance of DPA when facing various data augmentation techniques. Standard refers to the basic DPA with no modifications. This table evident that our method can still render models vulnerable under most data augmentation methods, indicating that our approach possesses strong robustness.



\begin{table}[t]
\centering
\begin{tabular}{ccccc}
\hline
Method     & ACC  & $\hat{\rho}_{F}\downarrow$ & $\hat{\rho}_{P}\downarrow$ & $\hat{\rho}_{D}\downarrow$             \\ \hline
Clean      & 0.81 & 1.62          & 1.01          & 1.24          \\ \hdashline
Standard   & 0.76 & 0.79          & 0.57          & 0.39          \\
Pretrained & 0.75 & 1.12          & 0.59          & 0.42          \\
Cutout     & 0.77 & 0.92          & 0.60          & 0.44          \\
Gaussian   & 0.78 & 0.87          & 0.58          & 0.42          \\
Gray       & 0.73 & 0.62 & 0.53 & 0.35 \\
JPEG       & 0.79 & 1.16          & 0.64          & 0.60          \\ \hline
\end{tabular}
\caption{The accuracy and robustness(\%) of poisoned ResNet18 over CIFAR10 after data augmentation.}
\label{tab:defense performence}
\end{table}

\subsection{Adversarial training} Table \ref{tab:defense} demonstrates the effectiveness of DPA attacks against various adversarial training methods, where AT denotes adversarial training using PGD and SAM, Smoothing, TRADES are proposed by \cite{foret2021sharpnessaware, cohen2019certified, zhang2019theoretically} respectively. As shown in the table, traditional adversarial training methods are not entirely effective in defending against our attacks. Consequently, we propose a new training paradigm that directly decreasing the curvature of the loss function with respect to the input(namely, \textbf{Ours}). 

\begin{table}[t]
\centering
\begin{tabular}{ccccc}
\hline
Method     & ACC  & $\hat{\rho}_{F}\uparrow$ & $\hat{\rho}_{P}\uparrow$ & $\hat{\rho}_{D}\uparrow$             \\ \hline
Clean      & 0.81 & 1.62          & 1.01          & 1.24          \\
No defense & 0.76 & 0.79          & 0.57          & 0.39          \\
SAM        & 0.79 & 0.84          & 0.57          & 0.41          \\
AT         & 0.79 & 1.07          & 0.78          & 0.79          \\
Smoothing & 0.75 & 1.11 & 0.80 & 0.92
\\
TRADES & 0.78 & 1.19 & 0.82 & 0.85
\\
Ours       & 0.71 & \textbf{1.93} & \textbf{1.31} & \textbf{1.54} \\ \hline
\end{tabular}
\caption{The comparison of the accuracy and robustness(\%) of different adversarial training methods against DPA.}
\label{tab:defense}
\end{table}

This finding warns researchers that simply adopting commonly used adversarial training strategies is not enough to ensure a trustworthy model. Considering the curvature information maliciously used by our DPA is also crucial. Like most pioneering works in adversarial machine learning, we present a novel attack method, and a defense method naturally follows to design a more robust model.

\section{Conclusion}
In this paper, we introduce the DPA, a stealthy threat in AI that performs normally during training and validation. DPA targets the Hessian matrix, inducing a rapid convergence to a steep local optimum on poisoned data, thereby diminishing the model's robustness. We demonstrate DPA's performance through comparative adversarial analyses with traditional attacks. Notably, DPA demands lower costs and exhibits transferability. We also establish its resilience against common augmentations and pretrained defenses, highlighting its universal risk underscores a novel challenge for AI security.

\section{Acknowledgments}
This research was partially supported by the National Natural Science Foundation of China (Grant Nos. 62202009 and 62571284), the
Macau Science and Technology Development Fund (Grant
Nos. 0040/2023/ITP1 and 0004/2023/RIB1), and the Basic and Applied Basic Research Foundation of Guangdong
Province (Grant No. 2024A1515011755).

\bibliography{aaai2026}

\appendix
\section{Appendix}

\section{Complexity analysis}

Here is a specific analysis of computational complexity:

\paragraph{Time complexity} Assuming the model has a total of $M$ parameters and the input $x$ has dimensions (3, 224, 224), denoted by 
$P=3\times224\times224$. The estimated theoretical time complexity of our proposed method is as follows:

\begin{itemize}
    \item Stage 1. Train Model: Update model parameters: \(O(MP)\)  \cite{shah2022time}
    \item Stage 2. Update perturbation
    \begin{itemize}
        \item Step 1. Perform a forward pass to obtain the Loss: \(O(MP)\)
        \item Step 2. Calculate the Jacobian matrix of the Loss with respect to \(x+\delta\): \(O(P)\)
        \item Step 3. Use the HVP (Hessian-vector Product) to estimate \(Hv\): \(O(P)\) (Or directly calculate the Hessian: \(O(P^2)\))
        \item Step 4. Compute \(\text{tr}(H^TH)\) or \(\|Hv\|_F\): \(O(P)\)
        \item Step 5. Backpropagate and update \(\epsilon\): \(O(P)\)
    \end{itemize}
\end{itemize}

In summary, generating perturbations using the original method has an overall time complexity of $O(MP)+O(P^2)+O(P)$, while using HVP reduces the time complexity to $O(MP)+O(P)$.

\paragraph{Space complexity} Since the space complexity of our proposed algorithm is independent of the model, we assume that the space complexity of Stage 1 is fixed at $O_1$. We will primarily analyze the space complexity of the Hessian part.

After obtaining the Jacobian matrix of the loss function with respect to $X+\delta$, whose space complexity is $O(P)$, the original method computes the gradient of each element of the Jacobian matrix with respect to $X+\delta$, resulting in a computational complexity of $O(P^2)$. However, HVP first uses a vector $v$ as a weight matrix to compute a weighted sum of the Jacobian matrix, then uses PyTorch's computation graph to perform backpropagation on $X+\delta$ to approximate $Hv$, this process maintains a space complexity of $O(P)$.

In summary, we reduce the space complexity of the Hessian part from $O(P^2)$ to $O(P)$. Specifically, for ImageNet images, generating perturbations requires 13,842 MiB of GPU memory.

\section{Experiment}

\subsection{Experiment details}
\label{apd:details}
Table \ref{tab:Hyperparameter} presents the hyperparameter settings relevant to the generation of DPA perturbations in our experiments.

\begin{table}[h]
\centering
\resizebox{0.45\textwidth}{!}{
\begin{tabular}{ccc}
\hline
Parameter            & Value & Explanation                             \\ \hline
seed                 & 0     & Random seed.                            \\
lr                   & 0.01  & Learning rate of model training.        \\
num\_epochs          & 20    & Iterations of model training.           \\
batch\_size          & 64    & The batch size of training data.        \\
optimizer            & SGD   & The optimizer of model training.        \\
optimizer\_epsilon   & Adam  & The optimizer of perturbation updating.        \\
lr\_epsilon          & 0.001 & Learning rate of perturbation updating. \\
num\_epochs\_epsilon & 20    & Iterations of perturbation updating.    \\ \hline
\end{tabular}%
}
\caption{Hyperparameter Settings for Generating Perturbations.}
\label{tab:Hyperparameter}
\end{table}

\begin{figure*}[t]
    \centering
    \includegraphics[width=0.8\textwidth]{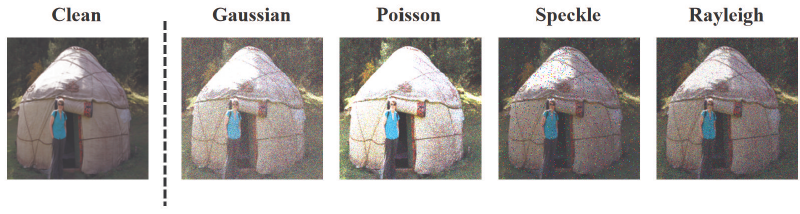}
    \caption{Image after adding different random noise}
    \label{fig：realworld}
\end{figure*}

\begin{table*}[t]
\centering

\begin{tabular}{c|cccc|cccc|cccc}
\hline
\multirow{2}{*}{Method} & \multicolumn{4}{c|}{VGG16}   & \multicolumn{4}{c|}{ResNet18} & \multicolumn{4}{c}{ResNet50} \\ \cline{2-13} 
                        & ACC  & $\hat{\rho}_{F}\downarrow$ & $\hat{\rho}_{P}\downarrow$ & $\hat{\rho}_{D}\downarrow$ & ACC  & $\hat{\rho}_{F}\downarrow$ & $\hat{\rho}_{P}\downarrow$ & $\hat{\rho}_{D}\downarrow$ & ACC  & $\hat{\rho}_{F}\downarrow$ & $\hat{\rho}_{P}\downarrow$ & $\hat{\rho}_{D}\downarrow$ \\ \hline
Clean                   & 0.89 &1.86&0.94&1.19& 0.81 &1.62&1.01&1.24& 0.78 &2.84&1.48&1.21\\
\hdashline
Standard                & 0.75 &0.58&0.50&\textbf{0.32}& 0.76 &0.79&0.57&0.39& 0.73 &1.38&0.62&0.45\\
Pretrained              & 0.78  & 0.85 & 0.53&0.35& 0.75  & 1.12 & 0.59&0.42& 0.72  & \textbf{0.66} &\textbf{0.55} & \textbf{0.34} \\
Cutout                  & 0.77  & 0.61 & 0.50&0.34& 0.77  & 0.92& 0.60&0.44& 0.75  & 1.45 & 0.59& 0.41 \\
Gaussian                & 0.75  & \textbf{0.55} & \textbf{0.49}&\textbf{0.32}& 0.78  & 0.87 & 0.58&0.42& 0.74  & 1.39 & 0.63& 0.41 \\
Gray                    & 0.72  & 0.60 & \textbf{0.49}&\textbf{0.32}& 0.73  & \textbf{0.62} & \textbf{0.53}&\textbf{0.35}& 0.69  & 0.82& 0.58& 0.35   \\
JPEG                    & 0.78  & 0.85 & 0.58 &0.45 & 0.79  & 1.16 & 0.64 & 0.60 & 0.78  & 0.98& 0.60 & 0.49   \\

\hline
\end{tabular}
\caption{The accuracy and robustness(\%) of poisoned models over CIFAR10 after data augmentation.}
\label{tab:defense performence}
\end{table*}

\subsection{Visualization of poisoned samples}

As depicted in Fig. \ref{fig：perturbation_compare}, the perturbation we propose is difficult to discern with the naked eye. Compared to other poisoning and backdoor attacks, even when the residual images between the poisoned and clean images are magnified five times, the noise is hardly noticeable. This illustrates the profound visual stealthiness inherent in our method, which is alarming in terms of its potential for misuse.

\subsection{Visualization of nature noise}
\label{apd:Visualization of Nature Noise}

We experimented with natural noise from various distributions to simulate real-world scenarios as show in Fig. \ref{fig：realworld}:
\begin{itemize}
    \item \textbf{Gaussian Noise:} Gaussian Noise follows a normal distribution and is the most common type of noise found in the real world;
    \item \textbf{Poisson Noise:} Poisson Noise is a type of discrete noise that can simulate the imaging process under low-light conditions;
    \item \textbf{Speckle Noise:} Speckle Noise is typically related to the local brightness of an image and can simulate bright or dark spots that may appear in an image;
    \item \textbf{Rayleigh Noise:} Rayleigh Noise can simulate image degradation caused by atmospheric scattering or other similar effects, typically manifesting as fluctuations in image brightness, particularly in the edge regions of the image;
\end{itemize}

\subsection{Robustness}

Table \ref{tab:defense performence} completely displays the performance of deferred poisoning attacks when facing various data augmentation techniques with $\epsilon$ set to $3/255$. It is evident that our method can still render models vulnerable under most data augmentation methods as well as with pre-trained models, indicating that our approach possesses strong robustness.

\subsection{Different poisoning percentages} In real-world scenarios, it is highly probable that attackers do not have complete access to all data. More challengingly, we simulate the hazards of deferred poisoning attacks under various realistic conditions using different percentages of poisoned data.

\begin{table}[t]
\centering

\begin{tabular}{ccccccc}
\hline
Robustness & 0\%  & 20\% & 40\% & 60\% & 80\% & 100\%         \\ \hline
ACC     & 0.89 & 0.88 & 0.87 & 0.86 & 0.83 & 0.75 \\
$\hat{\rho}_{F}\downarrow$       & 1.86 & 2.53 & 1.35 & 1.06 & 1.08 & \textbf{0.58} \\
$\hat{\rho}_{P}\downarrow$       & 0.94 & 0.97 & 0.71 & 0.65 & 0.63 & \textbf{0.50} \\
$\hat{\rho}_{D}\downarrow$       & 1.19 & 1.03 & 0.73 & 0.65 & 0.56 & \textbf{0.32} \\ \hline
\end{tabular}%
\caption{The accuracy and robustness(\%) of VGG16 under different poisoning percentages in CIFAR10.}
\label{tab:percentage}
\end{table}

Table \ref{tab:percentage} displays the performance of the VGG16 model under various percentages of deferred poisoning attacks on the CIFAR10 dataset. As the poisoning ratio increases, the model becomes increasingly vulnerable. When the poisoning ratio reaches $40\%$, the deferred poisoning attack has already demonstrated its offensiveness. Even at a ratio of $60\%$, the model suffers significant harm, which further illustrates the severe threat that deferred poisoning attacks pose to machine learning, poisoning only a portion of the training data can pose a serious risk to the model.

\subsection{Adversarial training} Table \ref{tab:defense} completely demonstrates the effectiveness of DPA attacks against various adversarial training methods, where AT denotes adversarial training using PGD \cite{PGD}. As shown in the 6th row, traditional adversarial training methods are not entirely effective in defending against our attacks. For instance, when training CIFAR10 with VGG16 and under the FGSM attack, the robustness of the clean model is 1.86\%, whereas the robustness of the model poisoned after adversarial training is only 0.71\%. 

\begin{table*}[t]
\centering

\renewcommand{\arraystretch}{1.2}  
\begin{tabular}{c|cccc|cccc|cccc}
\hline
\multirow{2}{*}{} & ACC   & $\hat{\rho}_{F}\uparrow$    & $\hat{\rho}_{P}\uparrow$    & $\hat{\rho}_{D}\uparrow$    & ACC    & $\hat{\rho}_{F}\uparrow$    & $\hat{\rho}_{P}\uparrow$    & $\hat{\rho}_{D}\uparrow$    & ACC    & $\hat{\rho}_{F}\uparrow$    & $\hat{\rho}_{P}\uparrow$    & $\hat{\rho}_{D}\uparrow$   \\ \cline{2-13} 
                  & \multicolumn{4}{c|}{VGG16-CIFAR10} & \multicolumn{4}{c|}{ResNet18-CIFAR10} & \multicolumn{4}{c}{ResNet50-CIFAR10} \\ \hline
Clean             & 0.89 & 1.86 & 0.94 & 1.19 & 0.81  & 1.62  & 1.01  & 1.24 & 0.78  & 2.84  & 1.48 & 1.21 \\
No defense        & 0.75 & 0.58 & 0.50 & 0.32 & 0.76  & 0.79  & 0.57  & 0.39 & 0.74  & 1.38  & 0.62 & 0.45 \\
SAM               & 0.76 & 0.57 & 0.50 & 0.30 & 0.79  & 0.84  & 0.57  & 0.41 & 0.78  & 0.98  & 0.57 & 0.39 \\
AT                & 0.78 & 0.71 & 0.77 & 0.94 & 0.79  & 1.07  & 0.78  & 0.79 & 0.78  & 1.51  & 0.86 & 0.87 \\
Ours              & 0.80 & \textbf{1.72} & \textbf{1.25} & \textbf{1.52} & 0.71  & \textbf{1.93}  & \textbf{1.31}  & \textbf{1.54} & 0.68  & \textbf{2.51}  & \textbf{1.54} & \textbf{1.83} \\ \hline
\end{tabular}
\caption{The comparison of the accuracy and robustness(\%) of different kinds of adversarial training method against DPA.}
\label{tab:defense}
\end{table*}

\begin{table*}[ht]
\centering
\begin{tabular}{cc|cccc|cccc}
\hline
\multicolumn{2}{c|}{\multirow{2}{*}{}}                 & \multicolumn{4}{c|}{CIFAR10} & \multicolumn{4}{c}{SVHN}   \\ \cline{3-10} 
\multicolumn{2}{c|}{}          & ACC  & $\hat{\rho}_{F}\downarrow$    & $\hat{\rho}_{P}\downarrow$    & $\hat{\rho}_{D}\downarrow$    & ACC   & $\hat{\rho}_{F}\downarrow$    & $\hat{\rho}_{P}\downarrow$    & $\hat{\rho}_{D}\downarrow$    \\ \hline
\multicolumn{1}{c|}{\multirow{3}{*}{VGG16}}    & Clean & 0.90  & 1.86  & 0.94 & 1.19 & 0.95 & 3.98 & 1.98 & 5.39 \\
\multicolumn{1}{c|}{} & Origin & 0.72 & 0.54 & 0.45 & 0.30 & 0.83 & 1.42 & 0.86 & 1.04 \\
\multicolumn{1}{c|}{} & HVP    & 0.75 & 0.58 & 0.50 & 0.32 & 0.89 & 1.54 & 0.93 & 1.10 \\ \hline
\multicolumn{1}{c|}{\multirow{3}{*}{ResNet18}} & Clean & 0.81  & 1.62  & 1.01 & 1.24 & 0.94  & 3.33 & 1.73 & 2.57 \\
\multicolumn{1}{c|}{} & Origin & 0.74 & 0.72 & 0.49 & 0.31 & 0.84 & 1.65 & 0.96 & 1.12 \\
\multicolumn{1}{c|}{} & HVP    & 0.76 & 0.79 & 0.57 & 0.39 & 0.89 & 1.78 & 1.03 & 1.24 \\ \hline
\multicolumn{1}{c|}{\multirow{3}{*}{ResNet50}} & Clean & 0.79  & 2.84  & 1.48 & 1.21 & 0.92 & 3.24 & 1.59 & 2.48 \\
\multicolumn{1}{c|}{} & Origin & 0.70 & 1.28 & 0.51 & 0.38 & 0.84 & 1.54 & 0.93 & 1.12 \\
\multicolumn{1}{c|}{} & HVP    & 0.73 & 1.38 & 0.62 & 0.45 & 0.88 & 1.63 & 0.97 & 1.16 \\ \hline
\end{tabular}
\caption{The comparison of the robustness(\%) compared with the origin method and HVP.}
\label{tab:hvp}
\end{table*}

The results also show that SAM is ineffective against our attack method. For example, the model robustness of VGG16-CIFAR10, with and without SAM adversarial training, is nearly the same, 0.57\% (the 5th row of the 3rd column) and 0.58\% (the 4th row of the 3rd column), respectively. SAM is ineffective possibly because it focuses on decreasing the curvature of the loss function with respect to model parameters. In contrast, our method aims to increase the curvature of the loss function concerning the input examples. They are two different dimensionalities of the loss function. Therefore, SAM cannot effectively counter our attack method.

Consequently, we propose a new training paradigm that directly decreasing the curvature of the loss function with respect to the input(namely, \textbf{Ours} in Table \ref{tab:defense}). For example, in the 4th row and the 7th row of the table, the robustness increases from 0.58\% to 1.72\%.  This finding warns researchers that simply adopting commonly used adversarial training strategies is not enough to ensure a trustworthy model. Considering the curvature information maliciously used by our DPA is also crucial. In summary, our work focuses on revealing a new attack. Like most pioneering works (e.g., FGSM, PGD) in adversarial machine learning, we present a novel attack method, and a defense method naturally follows to design a more robust model.

\subsection{Hessian-vector product performance} In the paper, we introduce the Hessian-vector product (HVP) to calculate the relaxed solution of the Hessian singularization term more quickly. With the new strategy, it now takes about 240 minutes to poison the entire CIFAR10, and about 30 hours to Tiny-ImageNet for large-scale scenarios, which is comparable to the competitive methods in our paper as shown in Table \ref{tab:hvp}.

\begin{figure*}[t]
    \centering
    \includegraphics[width=0.8\textwidth]{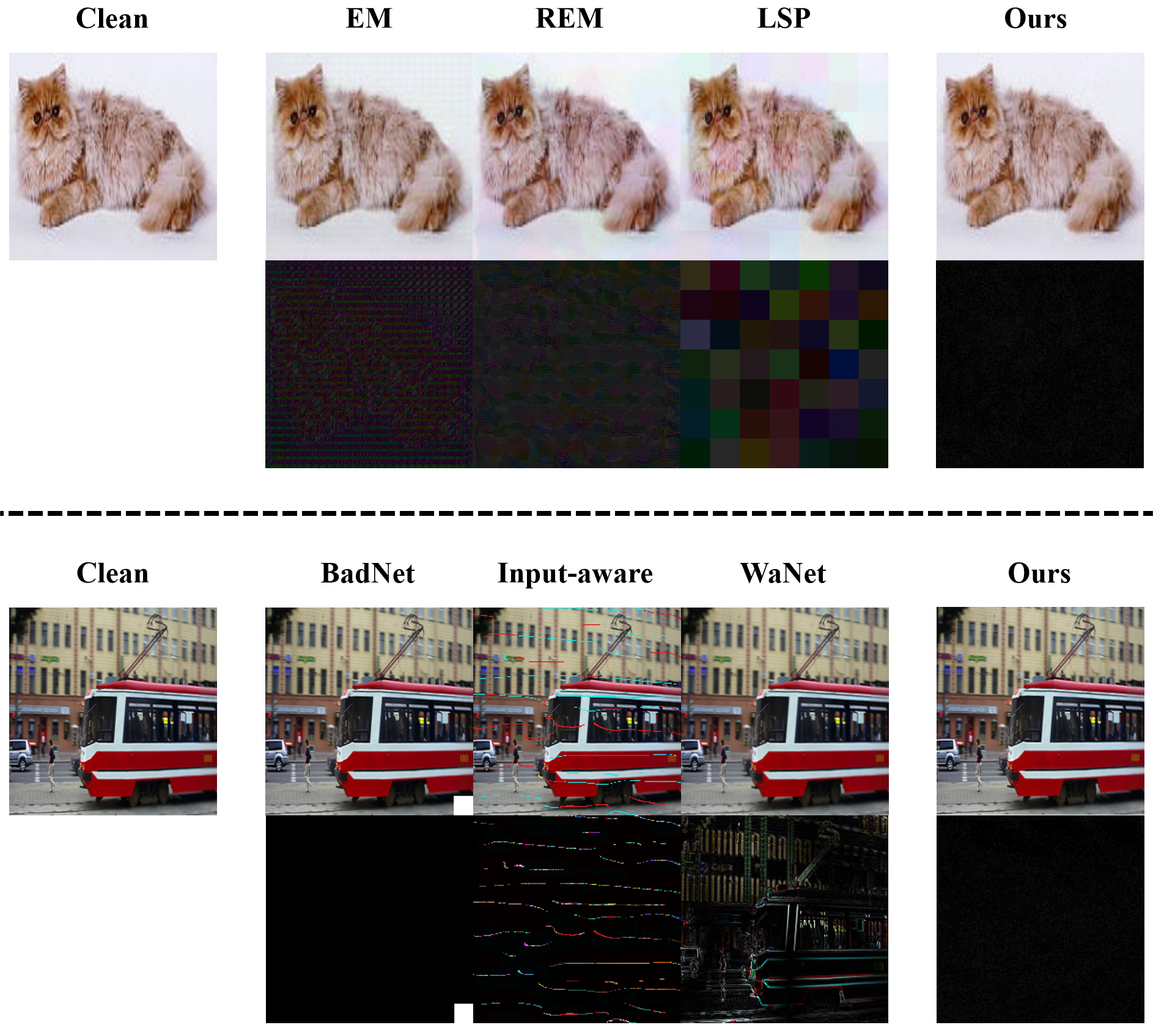}
    \caption{Comparison of examples generated by different Poisoning attacks above the dotted line(i.e. AR, EM, REM and LSP) and Backdoor attack below the dotted line(i.e. BadNet, Input-aware, LC and WaNet). For each attack, we show the poisoned sample (top) and the magnified (×5) residual (bottom).}
    \label{fig：perturbation_compare}
\end{figure*}

For example, as shown in the 1st-3rd row of the 1st column of Table \ref{tab:hvp}, when training CIFAR10 with VGG16 after being attacked by the original DPA, the model's robustness under FGSM attack drops from 1.86\% to 0.54\%. In contrast, when training with the dataset under HVP rapid attack, the robustness is 0.58\%, which is close to the original attack method. This indicates that HVP can accelerate the poisoning process without compromising the effectiveness of the DPA attack.

\end{document}